\documentclass{bmvc2k}

\title{DoublesEval: Diagnosing Multi-Agent Tactical Reasoning in Vision-Language Models via Professional Doubles Badminton}
\usepackage{booktabs}
\addauthor{Jintao Cheng}{jchengau@connect.ust.hk}{1}
\addauthor{Weibin Li}{yc67383@um.edu.mo}{2}

\addinstitution{
 The Hong Kong University of Science and Technology\\
 Hong Kong, China
}
\addinstitution{
 University of Macau\\
 Macau, China
}

\runninghead{Cheng and Li}{DoublesEval}

\begin{document}

\maketitle

\begin{abstract}
Visual Language Models (VLMs) excel at describing visible scene content but struggle to reason about dynamic multi-agent interactions, where action semantics depend on coordinated roles and spatial-temporal dependencies. We formalize this capability as \textbf{multi-agent tactical reasoning} and introduce \textbf{DoublesEval}, a diagnostic evaluation framework that leverages professional doubles badminton as a structurally tractable testbed. DoublesEval employs a key-moment-based protocol that decomposes rallies into tactically salient instants and probes models across four interpretable dimensions: atomic recognition, intra-segment composite understanding, cross-segment causal reasoning, and high-level tactical abstraction. This design isolates \emph{where} reasoning fails, rather than merely measuring answer correctness. To address observed failure modes, we propose \textbf{TacticCheck}, a lightweight constraint-guided test-time consistency checker that reranks candidate answers using the model's own lower-level tactical predictions, requiring no parameter updates or ground-truth labels at inference time. Evaluating four representative open-source VLMs on 60 curated rallies (yielding $\sim$9.6K structured instances) via a zero-shot protocol, we find that models remain weak across all diagnostic levels, with especially clear bottlenecks in spatial state, interaction binding, and terminal evidence. TacticCheck delivers consistent gains across all evaluated models, while still leaving a substantial gap to robust tactical reasoning. These results highlight the need for structured, interaction-aware evaluation paradigms for next-generation VLMs. The source code is available in \href{https://github.com/Chengjt1999/DoublesEval}{\textcolor{blue}{our GitHub repository}}.
\end{abstract}


\section{Introduction}

\begin{figure}[t]
    \centering
    \includegraphics[width=\linewidth]{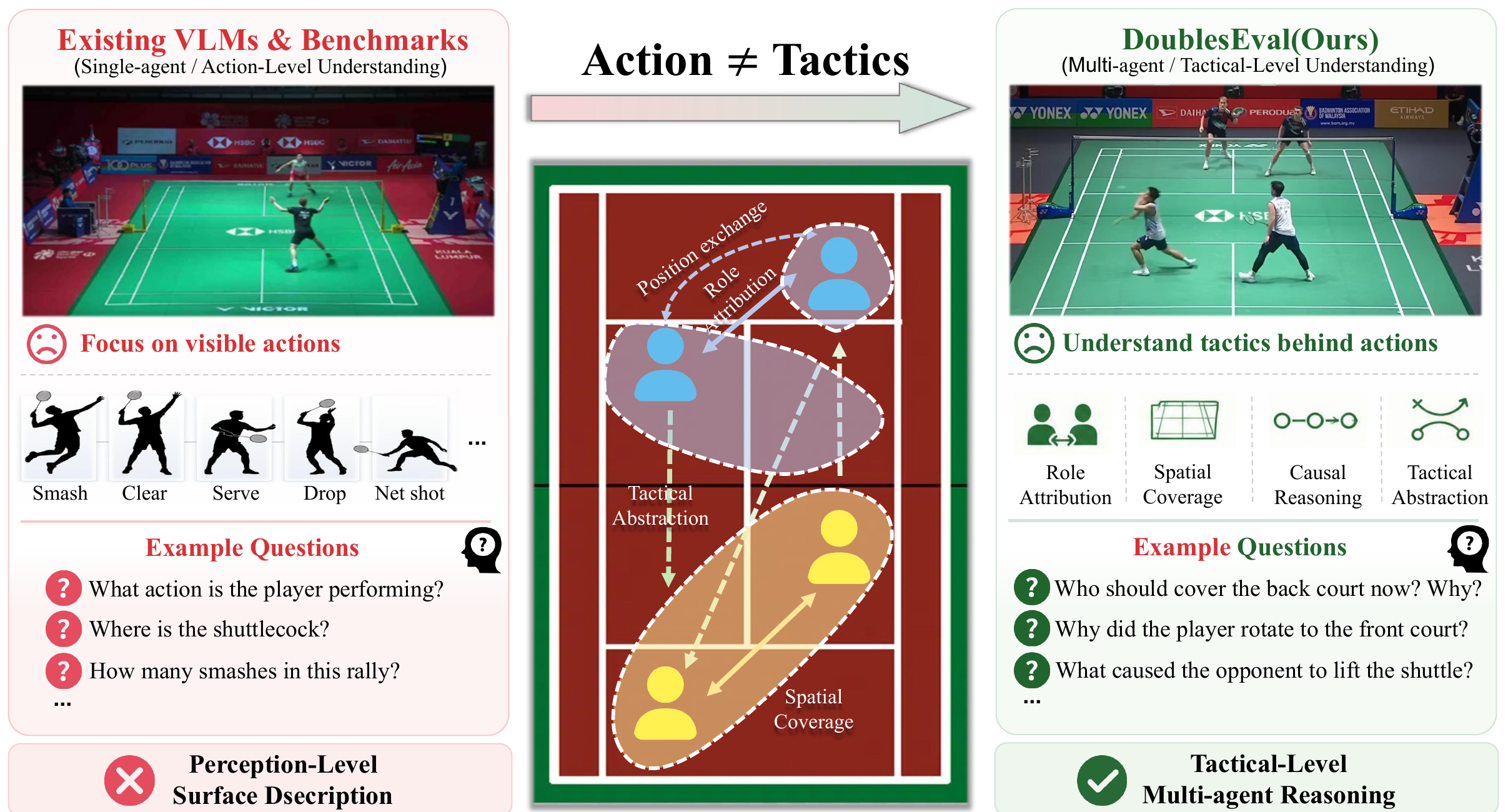}
    \caption{Comparison between existing action-level evaluations and our proposed \textbf{DoublesEval} framework. \textbf{Left:} Current VLM benchmarks primarily focus on perception-level surface descriptions, such as isolated action recognition (e.g., distinguishing a smash from a clear). \textbf{Middle:} The core semantic gap lies in multi-agent tactical coordination, which requires mapping atomic actions to dynamic role attributions, spatial coverage constraints, and causal chains. \textbf{Right:} DoublesEval addresses this by introducing a diagnostic framework tailored for tactical-level multi-agent reasoning, systematically probing models across four structured reasoning dimensions.}
    \label{fig:teaser}
\end{figure}

Recent advancements in Visual Language Models (VLMs) have substantially elevated performance on visual question answering, video captioning, and multimodal reasoning, largely attributable to large-scale vision-language pretraining and the integration of more capable temporal encoders~\cite{liu2024improved, lin2024videollava, cheng2025scale}. Despite these gains, model capabilities remain predominantly anchored to \emph{what is visible}---static objects, isolated scenes, or short-horizon actions---rather than \emph{what is happening between agents over time}.

Many real-world visual scenarios are fundamentally interactive: multiple agents move simultaneously, exchange roles, and coordinate their behavior, and the meaning of any single action depends on this collective context. We refer to this capability as \textbf{multi-agent tactical reasoning}---jointly modeling agent identity, spatial configuration, temporal evolution, and role-dependent interaction. Prevailing video evaluation protocols and datasets~\cite{kay2017kinetics, caba2015activitynet} are predominantly designed to assess \emph{individual} agent actions or \emph{global} event classifications. Consequently, they rarely stress-test a model's capacity for interactive, context-dependent reasoning, such as evaluating coordinated coverage or rotation necessity. As a result, recurring failure modes in role attribution, spatial coverage, and cross-event causal reasoning remain largely invisible under current evaluation protocols---even in recent domain-specific efforts like FineBadminton~\cite{he2025finebadminton}, which emphasizes fine-grained action recognition over the structured, multi-agent tactical coordination that recent analyses show VLMs struggle with~\cite{zhang2024vision, yuksekgonul2022and}.

We argue that \textbf{professional doubles badminton} offers a near-ideal testbed for studying this gap. It is, in effect, a \emph{structurally tractable} multi-agent tactical environment: only four players on a strictly bounded court, with well-defined team structure, rapid role transitions (attack/defense/rotation/coverage), and dense causal chains where one shot mechanically constrains the next. Unlike team sports such as football or basketball---where the agent count and spatial scale make ground-truth role labeling prohibitive~\cite{deliege2021soccernetv2}---doubles badminton is small enough to annotate exhaustively, yet rich enough to require genuine tactical understanding rather than surface-level description~\cite{he2025finebadminton}.

Building on this observation, we introduce \textbf{DoublesEval}, a diagnostic framework for evaluating zero-shot multi-agent tactical reasoning of VLMs on professional doubles badminton videos (see Figure~\ref{fig:teaser}). Instead of coarse video-level QA, DoublesEval adopts a \emph{key-moment-based} protocol: each rally is decomposed into tactically salient instants (e.g., serves, defensive transitions, and smash returns) and probed along four diagnostic dimensions: (1) \textit{single-field atomic recognition}, which examines basic perception of players, court regions, and actions; (2) \textit{intra-segment composite understanding}, which evaluates whether models can combine multiple visual cues within a local segment; (3) \textit{cross-segment causal reasoning}, which probes connections between earlier tactical decisions and later developments; and (4) \textit{high-level tactical semantic abstraction}, which assesses whether models can summarize broader tactical patterns. This decomposition lets us pinpoint \emph{where} reasoning breaks down, not merely whether the final answer is correct.

Motivated by the failure patterns surfaced by this diagnosis, we further propose \textbf{TacticCheck}, a lightweight constraint-guided test-time consistency checker. Instead of retraining the model or prompting it to generate a new explanation, TacticCheck uses the model's own lower-level predictions as a tactical checklist and reranks higher-level candidate answers by penalizing schema-level contradictions with this checklist. Mirroring how a human coach would sanity-check a tactical reading, the method requires no parameter updates and no ground-truth labels during inference, providing a direct stress test of whether explicit tactical consistency can mitigate multi-agent reasoning errors.

We instantiate DoublesEval on \textbf{60 curated professional doubles badminton rallies}, yielding approximately \textbf{9{,}600 structured evaluation instances} across the four diagnostic dimensions. To handle the open-ended nature of free-form VLM outputs, we adopt a two-stage evaluation protocol: four representative open-source VLMs---\textbf{Qwen2.5-VL}~\cite{wang2024qwen2}, \textbf{Qwen3-VL}~\cite{bai2025qwen3}, \textbf{VideoLLaMA3}~\cite{zhang2025videollama}, and \textbf{Molmo2}~\cite{clark2026molmo2}---are first prompted to generate zero-shot short answers across all diagnostic prompts. Subsequently, \textbf{GPT-5}~\cite{singh2025openai} is employed as a structured label-mapping judge to align these free-form responses to standardized evaluation categories, effectively mitigating lexical variance without introducing subjective scoring. This protocol reveals recurring failures in spatial state, interaction binding, score detail, and cross-segment causality---failure modes that conventional video evaluation protocols are not designed to localize. TacticCheck yields consistent gains across all evaluated models, indicating that explicit tactical consistency can meaningfully mitigate---though not eliminate---structural reasoning errors.

Our contributions are summarized as follows:
\begin{itemize}
    \item We \textbf{formalize multi-agent tactical reasoning} as a distinct evaluation target for VLMs and introduce \textbf{DoublesEval}, a key-moment-based diagnostic framework built from 60 professional doubles badminton rallies yielding $\sim$9.6K instances organized around four interpretable reasoning dimensions.
    \item We \textbf{systematically evaluate} representative open-source VLMs, exposing recurring failure modes in player role attribution, spatial localization, and cross-segment causal reasoning that prior video evaluation protocols do not surface.
    \item We \textbf{propose TacticCheck}, a constraint-guided test-time consistency checker that delivers consistent zero-shot improvements without parameter updates or ground-truth labels during inference.
\end{itemize}

\section{Related Work}

\subsection{Diagnostic Evaluation of Vision-Language Models}
While large-scale benchmarks have successfully tracked the upward trajectory of vision-language models (VLMs), aggregate metrics often obscure systematic reasoning failures. A growing line of work has therefore shifted toward diagnostic evaluations that isolate specific cognitive bottlenecks. Early investigations demonstrated that contrastive VLMs frequently degenerate into bag-of-words representations, exhibiting pronounced deficits in attribute binding and relational composition~\cite{yuksekgonul2022and, thrush2022winoground,luo2025overlapmamba}. Subsequent studies on instruction-tuned multimodal LLMs have further revealed that even state-of-the-art architectures struggle with fine-grained visual grounding~\cite{tong2024eyes} and elementary spatial or geometric reasoning that humans perform effortlessly~\cite{rahmanzadehgervi2024vision}. In the video domain, recent diagnostic suites such as TempCompass~\cite{liu2024tempcompass} and Video-MME~\cite{fu2025video} have effectively exposed vulnerabilities in temporal ordering, event localization, and long-horizon dependency modeling. Despite these advances, existing diagnostic protocols remain overwhelmingly anchored to \emph{single-agent} trajectories or \emph{global} event descriptors. They rarely interrogate whether a model can disentangle the interdependent dynamics of multiple agents---where the semantic validity of an action is contingent on partner roles, spatial coverage, and cross-temporal causality. DoublesEval addresses this blind spot by formalizing multi-agent tactical reasoning as a distinct diagnostic target.

\vspace{-10pt}

\subsection{Video Understanding in Badminton and Racket Sports}
Computational analysis of badminton has evolved in tandem with broader advances in sports video understanding. Foundational efforts primarily targeted low-level perception, focusing on shuttlecock and player tracking~\cite{huang2019tracknet,cheng2024mf,zeng2024mambamos} or stroke-level action classification from broadcast footage~\cite{ghosh2018towards,cheng2026beyond}. As annotation granularity improved, research shifted toward modeling \emph{tactical units}, decomposing rallies into structured sequences annotated with shot types, player coordinates, and point outcomes. The ShuttleSet series~\cite{wang2023shuttleset, wang2023shuttleset22} and the associated CoachAI Badminton Challenge~\cite{wang2024coachai,li2025mdmu} have been instrumental in this transition, though predominantly within the singles domain. More recent initiatives have begun to incorporate doubles-specific dynamics, expanding annotation coverage to include partner coordination and fine-grained tactical descriptors~\cite{he2025finebadminton}. Nevertheless, these resources are fundamentally optimized for \emph{supervised, task-specific pipelines}---such as stroke classification, trajectory forecasting, or win-probability estimation---rather than for probing the open-ended, zero-shot reasoning capabilities of contemporary VLMs. Crucially, the complex interplay of role switching, court-coverage division, and role-dependent causal chains in doubles play has not been leveraged as a structured diagnostic signal for multimodal models. By repurposing professional doubles rallies into a key-moment-based evaluation protocol, DoublesEval bridges this methodological divide, offering the first diagnostic framework tailored to multi-agent tactical reasoning in sports video.
\section{Methodology}
\begin{figure}[t]
    \centering
    \includegraphics[width=\linewidth]{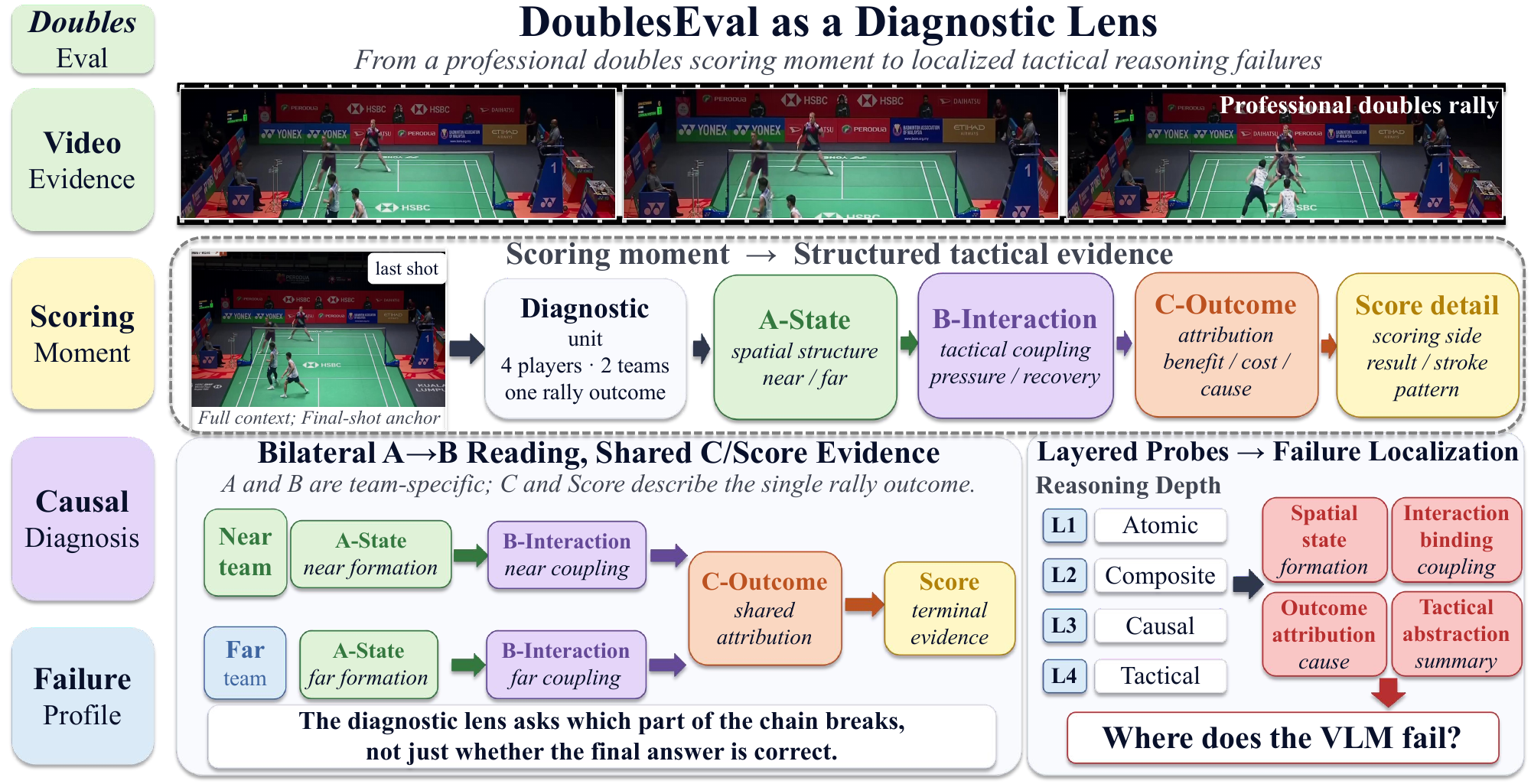}
    \caption{\textbf{Overview of the DoublesEval diagnostic pipeline.} By anchoring on the scoring moment, we extract structured tactical evidence across the bilateral causal chain. Layered reasoning probes (L1-L4) are then applied to precisely localize \emph{where} VLM multi-agent reasoning fails, rather than merely evaluating outcome correctness.}
    \label{fig:method_pipeline}
\end{figure}
\subsection{Task Formulation: Multi-Agent Tactical Reasoning}
\label{sec:task}
\paragraph{Formal Task Definition.}
As illustrated in Figure~\ref{fig:method_pipeline}, let $\mathcal{V}$ denote a professional doubles badminton rally video containing four players partitioned into two teams. The video is temporally segmented into an ordered sequence of \emph{key moments} $\mathcal{M} = \{m_1, m_2, \ldots, m_K\}$, where each $m_k$ corresponds to a tactically salient interval (e.g., serve, offensive transition, or defensive coverage). The multi-agent tactical reasoning task is formally defined as learning a conditional mapping $\mathcal{F}: (\mathcal{V}, m_k, q) \mapsto \hat{y}$, where $q$ is a natural-language query that conditions the model on a specific reasoning target, and $\hat{y} \in \mathcal{Y}$ is the predicted tactical inference. The target space $\mathcal{Y}$ encompasses agent role assignment, spatial occupancy, temporal causality, and strategic pattern recognition. Under a zero-shot setting, model capability is quantified by the alignment between $\hat{y}$ and the expert-annotated ground truth $y^\star$, with failures analyzed at the query level rather than aggregated video-level accuracy.

\paragraph{Diagnostic Dimension Stratification.}
To systematically isolate capability bottlenecks within $\mathcal{F}$, we partition the evaluation query space into four non-overlapping diagnostic dimensions, $\mathcal{D} = \{D_1, D_2, D_3, D_4\}$. This stratification is ordered by increasing reasoning depth and temporal scope, transforming a monolithic accuracy metric into a fine-grained diagnostic signal. Each dimension is operationally defined as follows:

\textbf{L1: Single-Field Atomic Recognition.} This dimension targets baseline visual grounding. Queries are constrained to a single key moment $m_k$ and require the model to identify isolated factual elements, such as player identity, court zone occupancy, or discrete stroke types. Success here verifies that the model's perceptual encoder can reliably resolve low-level visual signals without compositional interference.

\textbf{L2: Intra-Segment Composite Understanding.} This dimension probes multi-cue binding within a localized temporal segment. Queries require the simultaneous integration of co-occurring visual attributes (e.g., spatial relation + agent role + action state) to resolve interdependent facts. Failure at this stage typically indicates an inability to jointly model spatial configurations and agent states, even when temporal context is fixed.

\textbf{L3: Cross-Segment Causal Reasoning.} This dimension tests cross-temporal dependency modeling across $\geq 2$ sequential moments. Queries explicitly link a prior tactical decision (cause) to a subsequent player repositioning or stroke selection (effect). This isolates the model's capacity to construct implicit causal chains and maintain state continuity over dynamic multi-agent interactions.

\textbf{L4: High-Level Tactical Semantic Abstraction.} This dimension evaluates holistic pattern induction over extended rally phases. Queries require summarizing recurrent strategic configurations (e.g., rotation schemes, coverage division rules, or offensive/defensive transitions) rather than discrete events. Performance here reflects the model's ability to distill granular observations into structured tactical priors.

By attributing model failures to a specific dimension $D_i$, the protocol explicitly decouples perceptual grounding errors from higher-order reasoning breakdowns. This structured decomposition ensures that evaluation outcomes are interpretable, reproducible, and directly actionable for diagnosing the specific cognitive bottlenecks of modern VLMs.
\subsection{DoublesEval: A Diagnostic Suite for Tactical Reasoning}   
\subsubsection{Video Source and Rally Curation}
\paragraph{Source matches.}
DoublesEval is curated from four professional doubles matches selected from the BWF World Tour \emph{Super 750 and Super 1000} series, comprising two men's doubles, one women's doubles, and one mixed doubles match. All broadcast footage is sourced from the official \textbf{BWF TV} YouTube channel at 1080p resolution and 25/30 fps. 

\paragraph{Rally selection and tactical framing.}
From these four matches, we curate 60 rallies. Selection is guided by three criteria: (i) \emph{tactical diversity}, ensuring coverage of attacking, defending, rotation, and net-play situations; (ii) \emph{minimum tactical depth}, excluding single- and double-stroke rallies so that each example exercises the four diagnostic dimensions; and (iii) \emph{visual clarity}, excluding rallies with camera cuts, replays, or substantial occlusion. Crucially, we treat each rally $V$ as a holistic tactical unit: VLMs receive the \emph{full rally video} as temporal context to capture build-up and transition patterns, while diagnostic queries and expert annotations are anchored to \emph{tactically decisive segments} (e.g., scoring windows, critical rotations, or phase transitions). This design ensures that evaluation targets concrete tactical decisions without sacrificing the global strategic context required for D3/D4 reasoning. We deliberately prioritize \emph{tactical diversity per rally} over raw quantity, consistent with the diagnostic suite philosophy.

\paragraph{Rally statistics.}
The 60 curated rallies have an average duration of approximately $10$ seconds, with $\overline{S}$ strokes per rally on average. Each rally is partitioned into multiple diagnostic segments for fine-grained probing; per-rally segment density and full distributional statistics are reported in Table~\ref{tab:dataset_stats}.

\begin{table}[t]
\centering
\small
\renewcommand{\arraystretch}{1.15}
\caption{Dataset statistics for the 60 curated rallies. Diagnostic segments are tactically decisive intervals (e.g., scoring windows, role transitions) anchored within the full-rally context. Instance counts per dimension reflect balanced coverage across the four diagnostic targets. Annotation reliability metrics are computed on the 30\% expert-audited subset.}
\begin{tabular}{lccc}
\toprule
\textbf{Statistic} & \textbf{Mean} & \textbf{Std} & \textbf{Range} \\
\midrule
\multicolumn{4}{l}{\textit{Rally-level basics}} \\
Duration (seconds) & $10.2$ & $3.1$ & $[6.5, 18.9]$ \\
Strokes per rally ($\overline{S}$) & $12.4$ & $3.8$ & $[7, 23]$ \\
\midrule
\multicolumn{4}{l}{\textit{Diagnostic segment density}} \\
Segments per rally & $40.1$ & $5.3$ & $[32, 51]$ \\
Avg. segment duration (seconds) & $2.8$ & $0.6$ & $[1.8, 4.2]$ \\
\midrule
\multicolumn{4}{l}{\textit{Instance distribution by dimension}} \\
L1: Atomic Recognition & $2{,}400$ & ($25.0\%$) & -- \\
L2: Composite Understanding & $2{,}400$ & ($25.0\%$) & -- \\
L3: Causal Reasoning & $2{,}400$ & ($25.0\%$) & -- \\
L4: Tactical Abstraction & $2{,}400$ & ($25.0\%$) & -- \\
\midrule
\textbf{Total instances} & \multicolumn{3}{c}{\textbf{$9{,}600$}} \\
\midrule
\multicolumn{4}{l}{\textit{Annotation reliability (audit subset)}} \\
Intra-tier agreement ($\kappa$) & $0.78$--$0.86$ & -- & -- \\
Tier-to-expert alignment (Acc) & $82.5\%$--$89.5\%$ & -- & -- \\
\bottomrule
\end{tabular}

\label{tab:dataset_stats}
\end{table}

\paragraph{Coverage trade-off.}
We acknowledge that four source matches yield a comparatively small number of distinct player pairs and tactical styles. This is a deliberate design choice: a diagnostic suite that prioritizes dense expert annotation and per-dimension interpretability necessarily trades player-coverage breadth for annotation depth. To prevent trivial player-identity shortcuts, prompts that reference specific players are constructed using \emph{relative} role descriptors (e.g., ``the back-court player on the serving team'') rather than proper names; we further discuss this design in Section~\ref{sec:prompts}.

\subsubsection{Diagnostic Prompt Construction and Instance Design}
\label{sec:prompts}
\paragraph{Paired Query Architecture.}
The $\sim$9{,}600 evaluation instances are structured as paired diagnostic queries. For each rally $v$ and diagnostic dimension $D_i \in \{D_1, D_2, D_3, D_4\}$, we instantiate two complementary prompts targeting the tactical state within the scoring window $W_v$: a \emph{Single-Choice Question (SCQ)} and a \emph{Short-Answer Question (SAQ)}. This yields 4{,}800 SCQ instances and 4{,}800 SAQ instances. The SCQs provide a controlled evaluation baseline with exactly one correct option derived from the expert annotation schema, while the SAQs extend the same tactical context into open-ended generation to stress-test reasoning depth and justification capability without answer scaffolding.

\paragraph{SCQ Design and Construction.}
SCQ prompts are formulated to isolate specific perceptual or relational bottlenecks under strict single-choice constraints. Each question presents four mutually exclusive options: exactly one correct label sampled from the ground-truth annotation $\mathcal{L}_{v,d}$, and three distractors systematically constructed to reflect common model failure modes (e.g., swapped player roles, inverted court zones, or temporally misaligned causal links). This design ensures that SCQ accuracy measures model discrimination capability without ambiguity, serving as a reliable lower-bound for tactical comprehension in each dimension.

\paragraph{SAQ Design and Open-Ended Probing.}
SAQ prompts are derived by removing the single-choice options and appending a directive that requires the model to generate a concise tactical justification or descriptive summary (e.g., ``Explain why the back-court player rotated after the smash,'' or ``Describe the dominant coverage pattern during this phase.''). SAQs explicitly test the model's ability to articulate causal chains, bind multi-cue observations, and abstract strategic patterns without option scaffolding. By construction, SAQ outputs exhibit high lexical variance, making direct string-matching evaluation unreliable and necessitating semantic alignment.

\paragraph{Label-Mapping for SAQ Evaluation.}
To enable rigorous comparison with the SCQ baseline and ground-truth labels, all SAQ responses are processed through the structured label-mapping judge introduced in Stage 2. The judge receives the SAQ output alongside the dimension-specific label space $\mathcal{L}_{v,d}$ and maps the free-form text to the single semantically equivalent category. This step does not perform subjective scoring; it exclusively aligns open-ended reasoning to the standardized evaluation vocabulary established during annotation, ensuring that performance metrics reflect tactical comprehension rather than generative phrasing preferences. The paired SCQ/SAQ design allows us to quantify the ``reasoning gap'' between closed-set recognition and open-ended tactical articulation across all four dimensions.

\subsubsection{Expert-Verified Scoring-Window Annotation}
\label{sec:annotation}
\paragraph{Scoring window as annotation anchor.}
For each rally $V$, we define a \emph{scoring window} $W$ comprising the final scoring stroke and the 3--5 immediately preceding shots. While VLMs ingest the \emph{full rally video} as temporal context to capture long-horizon tactical evolution, expert annotations and diagnostic queries are exclusively anchored to the tactical state within $W$. This design mirrors professional coaching analysis: the scoring formation dictates the point outcome, but its tactical validity and intent are contingent on the immediate preceding sequence. Consequently, $W$ serves as the unified annotation target for all diagnostic dimensions, with L1/L2 probing the immediate state within $W$, and D3/D4 evaluating the causal and strategic links from earlier phases of $V$ to the configuration in $W$.

\paragraph{Two-tier annotation pipeline.}
To balance coverage breadth and tactical fidelity, annotation is organized into a structured two-tier pipeline. The \emph{first tier} consists of two experienced badminton analysts who independently annotate all 60 rallies. The \emph{second tier} comprises two certified \textbf{National Level-1 Athletes} (the highest non-professional competitive tier), who audit a stratified 30\% subsample ($\approx 18$ rallies) balanced across disciplines. Level-1 annotators adjudicate discrepancies, correct tactically implausible labels, and establish the gold standard for the audited subset. Their expertise ensures that complex role attributions and coverage patterns are grounded in domain-accurate reasoning rather than superficial visual cues.

\paragraph{Tool and schema.}
Annotation is conducted via \textbf{Label Studio}\cite{LabelStudio}, where experts assign dimension-specific tactical labels to the state within $W$. These labels directly populate the closed answer spaces $\mathcal{L}_{k,j}$ for MCQ construction and serve as semantic reference targets for SAQ label-mapping. To prevent trivial identity-based shortcuts, all player references in prompts and labels strictly use \emph{relative tactical descriptors} (e.g., ``the back-court player on the serving team'') rather than proper names.

\paragraph{Procedure and quality control.}
The two first-tier annotators label each rally independently. Label Studio automatically surfaces per-window label discrepancies for manual reconciliation. Within the 30\% audit subset, Level-1 reviewers override consensus labels when tactical reasoning is violated; this expert-corrected version serves as the gold standard. For the remaining 70\%, resolved consensus labels are used as working ground truth. Inter-annotator reliability on the audited subset shows substantial-to-almost-perfect intra-tier agreement (Cohen's $\kappa \in [0.80, 0.87]$ for perceptual labels, $\kappa \in [0.74, 0.82]$ for role attribution) and high tier-to-expert alignment ($>87\%$ accuracy for stroke intent, $>79\%$ for tactical role assignment). These metrics confirm that the tiered pipeline yields consistent, domain-validated annotations suitable for zero-shot diagnostic evaluation, while acknowledging the inherent subjectivity of tactical role attribution in dynamic doubles play.

\begin{figure}[t]
    \centering
    \includegraphics[width=0.8\linewidth]{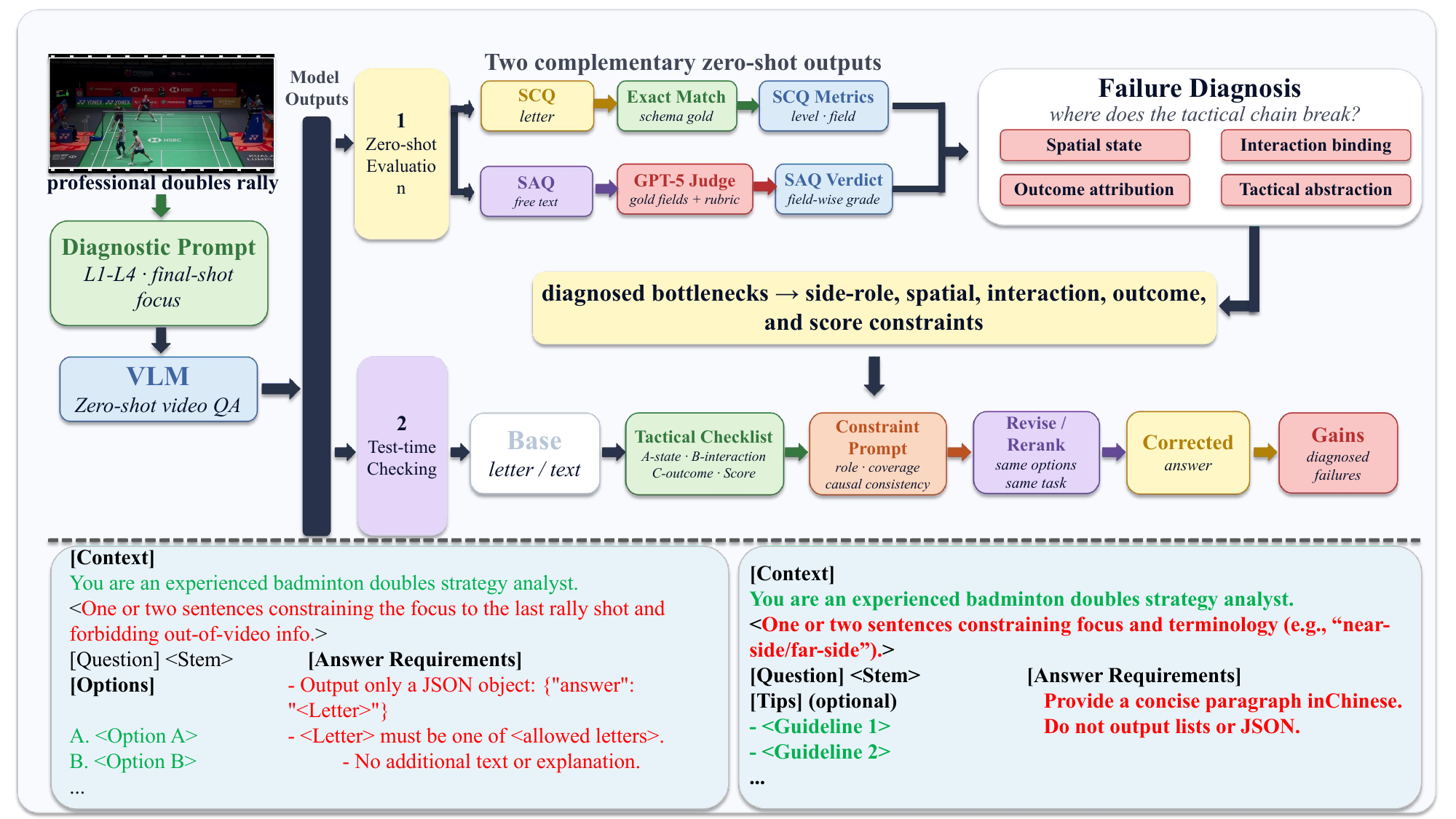}
    \caption{\textbf{Two-Stage Evaluation and TacticCheck Pipeline.} The framework processes key moments to generate two complementary zero-shot outputs: SCQs (evaluated via exact match) and SAQs (evaluated by a GPT-5 Judge). The diagnosed bottlenecks then inform a test-time checking loop, where a structured tactical checklist and constraint prompts guide the VLM to iteratively revise its initial answers.}
    \label{fig:eval_pipeline}
\end{figure}

\vspace{-8pt}

\subsection{Two-Stage Evaluation Protocol}
\subsubsection{Stage 1: Zero-Shot VLM Inference}
As illustrated in Figure~\ref{fig:eval_pipeline}, we evaluate four representative open-source VLMs: Qwen2.5-VL, Qwen3-VL, VideoLLaMA3, and Molmo2. Each model processes the full rally video $V_v$ as temporal context, paired with dimension-conditioned diagnostic prompts that explicitly target the tactical state within the scoring window $W_v$. Inference is conducted in a zero-shot setting with deterministic decoding (temperature $T=0.1$, top-$p=0.9$, max output tokens $=64$). The models produce free-form textual responses $A^{(0)}_{v,d}$ for each rally $v$ and diagnostic dimension $d \in \{D_1, D_2, D_3, D_4\}$.

\subsubsection{Stage 2: GPT-5 Label-Mapping Judge}
Open-ended VLM outputs exhibit substantial lexical variance, making direct string matching unreliable. To standardize evaluation, we employ GPT-5 as a structured label-mapping judge. The judge receives a fixed system prompt specifying: (i) the dimension-specific target label space $\mathcal{L}_{v,d}$ derived from the expert annotation of $W_v$, (ii) the VLM's free-form response $A^{(0)}_{v,d}$, and (iii) a strict output schema (JSON with fields \texttt{mapped\_label} and \texttt{confidence\_score}). GPT-5 performs semantic alignment rather than subjective scoring, mapping each open-ended response to the predefined evaluation vocabulary established during annotation. This step ensures consistent, reproducible label assignment across all evaluated models without introducing generative phrasing bias.
\subsubsection{Judge Reliability}
\label{sec:judge}
To validate the mapping fidelity of the LLM judge, we conduct a human-in-the-loop verification on a stratified random subset of $N_{\text{eval}} = 480$ instances (5\% of the full set). Two domain experts independently assign labels, and we compute inter-annotator agreement (Cohen's $\kappa = 0.86$) and judge-expert agreement ($\kappa = 0.81$). Discrepancies between the judge and experts are resolved via majority voting or expert arbitration. The judge demonstrates strong alignment with human annotations across all four dimensions, with the lowest agreement observed in \textit{Tactical Abstraction} ($\kappa = 0.73$), reflecting the inherent ambiguity of high-level strategic labeling. All downstream metrics are computed using the judge-mapped labels after reliability verification.

\subsection{TacticCheck: Constraint-Guided Test-Time Consistency Checking}
\label{sec:correction}
\subsubsection{Motivation}
The diagnostic analysis shows that many errors are not isolated wrong labels, but inconsistent tactical chains: a model may predict a plausible final outcome while assigning incompatible team formation, interaction role, or terminal stroke evidence. We therefore use the lower-level diagnostic fields as a test-time consistency signal. The goal is modest: TacticCheck does not teach the model new badminton knowledge, and it does not use ground-truth labels. It only checks whether a higher-level answer is compatible with the model's own tactical reading of the scoring moment.

\subsubsection{Checklist Extraction}
For each rally, we first collect the model's zero-shot predictions on L1 fields and convert them into a compact tactical checklist
\begin{equation}
    z = \{\hat{s}_{\text{near}}, \hat{s}_{\text{far}}, \hat{b}_{\text{near}}, \hat{b}_{\text{far}}, \hat{c}, \hat{r}\},
\end{equation}
where $\hat{s}_{\text{near}}$ and $\hat{s}_{\text{far}}$ denote near-team and far-team spatial states, $\hat{b}_{\text{near}}$ and $\hat{b}_{\text{far}}$ denote interaction cues, $\hat{c}$ denotes outcome attribution, and $\hat{r}$ denotes score-related terminal evidence. These fields are predicted by the same VLM under the same zero-shot setting. No expert annotation is used in this step.

\subsubsection{Constraint-Guided Option Reranking}
For a higher-level single-choice question with candidate options $\mathcal{O}=\{o_1,\ldots,o_N\}$, TacticCheck maps each option to a lightweight tactical signature $\phi(o_i)$ using the predefined answer schema. We then compute a consistency penalty between the option signature and the checklist:
\begin{equation}
    \mathrm{Penalty}(o_i,z)=
    \lambda_s \mathbf{1}[\phi_s(o_i) \not\sim z_s]
    + \lambda_b \mathbf{1}[\phi_b(o_i) \not\sim z_b]
    + \lambda_c \mathbf{1}[\phi_c(o_i) \not\sim z_c]
    + \lambda_r \mathbf{1}[\phi_r(o_i) \not\sim z_r],
\end{equation}
where $\not\sim$ indicates a schema-level incompatibility, such as a candidate requiring a far-team interaction when the checklist indicates a near-team interaction, or a score explanation that contradicts the predicted terminal evidence. The final answer is selected by reranking candidate options:
\begin{equation}
    \hat{o}=\arg\max_{o_i \in \mathcal{O}}
    \left[\mathrm{BaseScore}(o_i)-\mathrm{Penalty}(o_i,z)\right].
\end{equation}
In practice, $\mathrm{BaseScore}$ is derived from the model's original answer preference when available; otherwise the original selected option is retained as the default and only replaced when another option has a clearly lower consistency penalty. This conservative rule reduces unnecessary answer changes.

\subsubsection{Scope and Limitations}
TacticCheck is a lightweight test-time reranking method. It requires no parameter update, no additional training data, and no access to ground-truth labels during inference. Its benefit comes from down-ranking answers that contradict the model's own lower-level tactical evidence. However, the method is bounded by the quality of the checklist: if the L1 predictions are wrong, the consistency signal can also be wrong. We therefore treat TacticCheck as a diagnostic intervention that tests whether explicit tactical consistency helps, rather than as a complete solution to multi-agent tactical reasoning.

\subsection{Evaluation Metrics}
We treat every prompt instance as an evaluation unit after applying the GPT-5 label mapper. \textbf{Single-choice questions (SCQ)} are scored by exact matching the predicted option against the gold label. \textbf{Short-answer questions (SAQ)} are first mapped to the same discrete label space and then scored identically to SCQs. Dimension-wise accuracy $\text{Acc}_d$ averages the SCQ and SAQ accuracies within each level $d$ (the two formats appear in one-to-one pairs per rally), and the overall macro score is the unweighted mean across the four levels.

TacticCheck operates only on SCQs because the reranker requires an explicit option set. Its reported gains therefore compare the SCQ accuracy before reranking (``Base'') and after reranking on the identical SCQ subset; SAQ accuracy remains unchanged and is excluded from those deltas. Unless otherwise noted, tables that mix Base and TacticCheck results adopt this SCQ-only accounting for fair comparison. Statistical significance for the SCQ deltas is estimated via paired bootstrap over SCQ instances (1,000 resamples, $p<0.05$). Finally, failure-mode breakdowns are computed on the pre-reranking SCQ predictions so that error categories align with the structured distractor taxonomy.

\section{Experiments}
\label{sec:experiments}

\subsection{Overall Zero-Shot Performance}
\label{sec:overall_zeroshot}
\begin{figure}[t]
    \centering
    \includegraphics[width=0.8\linewidth]{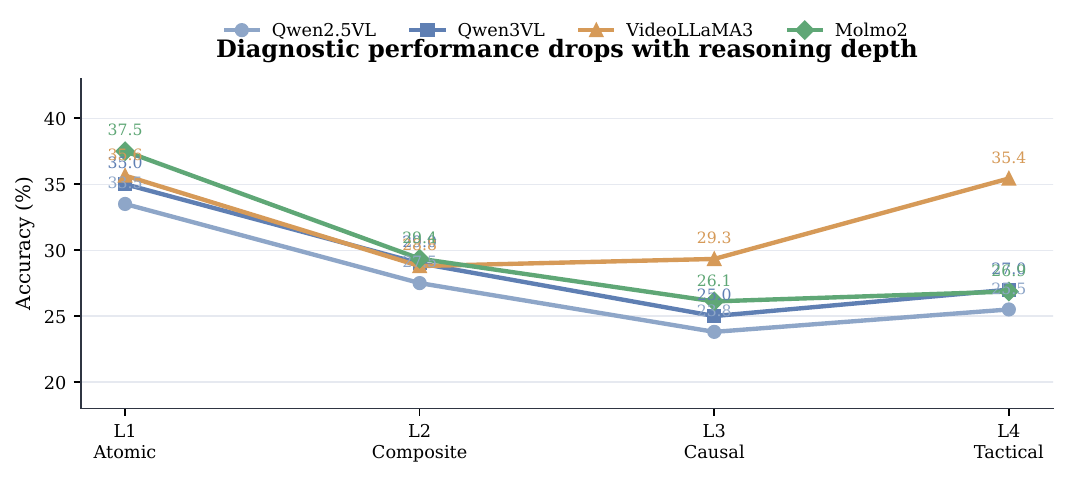}
    \caption{\textbf{Level-wise zero-shot performance over single-choice and short-answer questions.} The benchmark does not only test final answer correctness: it separates atomic recognition, composite understanding, causal reasoning, and tactical abstraction.}
    \label{fig:zeroshot_level}
\end{figure}

We first evaluate all models in a zero-shot setting. Each model receives the full rally video and answers diagnostic questions anchored to the scoring moment. Unless otherwise stated, level-wise results aggregate both single-choice questions and label-mapped short-answer questions. Table~\ref{tab:zeroshot_level} reports level-wise accuracy, and Figure~\ref{fig:zeroshot_level} visualizes the same trend.

\begin{table}[t]
\centering
\small
\setlength{\tabcolsep}{5pt}
\renewcommand{\arraystretch}{1.12}
\caption{\textbf{Zero-shot accuracy over single-choice and short-answer questions (\%).} Accuracy remains low across all four diagnostic levels. L1 is easier than the other levels for most models, but even atomic recognition is far from saturated.}
\begin{tabular}{lccccc}
\toprule
\textbf{Model} & \textbf{Overall} & \textbf{L1} & \textbf{L2} & \textbf{L3} & \textbf{L4} \\
\midrule
Qwen2.5VL   & 27.57 & 33.50 & 27.50 & 23.80 & 25.50 \\
Qwen3VL     & 29.00 & 35.00 & 29.00 & 25.00 & 27.00 \\
VideoLLaMA3 & 32.30 & 35.65 & 28.79 & 29.33 & 35.43 \\
Molmo2      & 31.31 & 37.50 & 29.35 & 26.11 & 26.86 \\
\bottomrule
\end{tabular}

\label{tab:zeroshot_level}
\end{table}

Overall performance falls in a narrow band, from $28.65\%$ to $32.35\%$. This indicates that current VLMs are not simply failing on one isolated model family. The weakness is shared across architectures under the same diagnostic protocol. Performance also does not form a perfectly monotonic curve over levels: for example, VideoLLaMA3 performs relatively well on L4. We therefore do not interpret the levels as a strict difficulty ladder for every model. Instead, the key observation is that all levels expose non-trivial errors, including the visually grounded L1 questions.

\subsection{Diagnostic Analysis}
\label{sec:diagnostic_analysis}
\begin{figure}[t]
    \centering
    \includegraphics[width=0.7\linewidth]{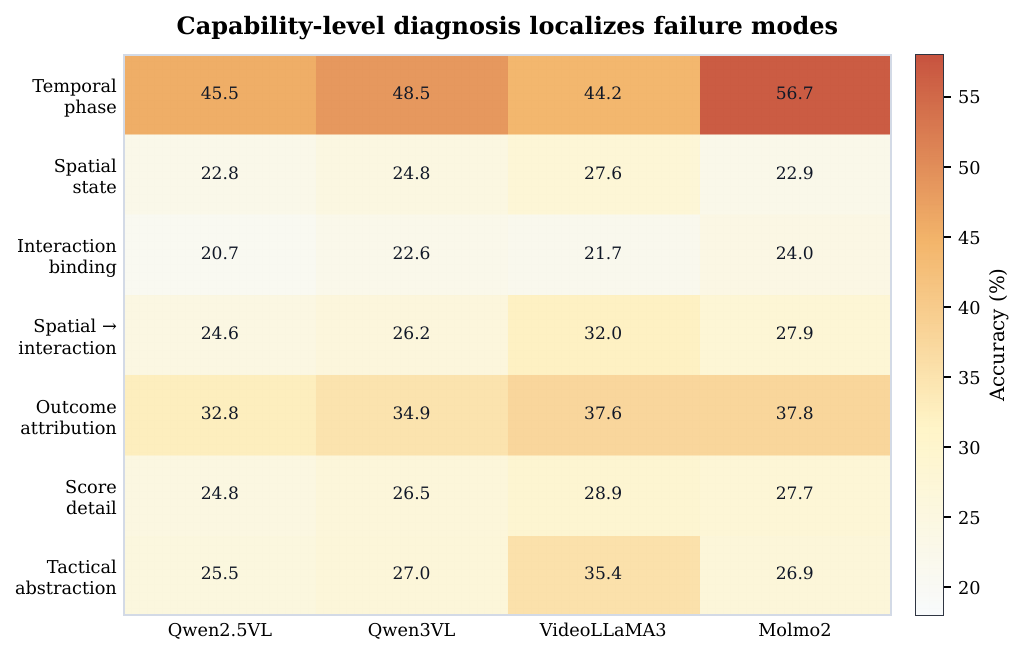}
    \caption{\textbf{Capability-level diagnosis.} Temporal phase is relatively easier, while spatial state, interaction binding, and score-detail reasoning are consistently weak. This suggests that models often see parts of the scene but fail to bind players, teams, space, and outcome into a coherent tactical state.}
    \label{fig:capability_heatmap}
\end{figure}

We next ask why zero-shot performance is low. DoublesEval is designed for this purpose: each question is tied to an interpretable capability, so errors can be localized rather than only counted. Figure~\ref{fig:capability_heatmap} summarizes accuracy by capability group.

\vspace{-5pt}
The main bottleneck is not basic visibility alone. Models achieve their best capability scores on temporal phase recognition, but drop on spatial state and interaction binding. These two categories require the model to identify who occupies which tactical role and how the two sides are coupled. Score detail is also difficult, especially when the answer depends on the final stroke pattern rather than only the winning side. This pattern supports the central motivation of DoublesEval: professional doubles badminton is compact enough to observe, but still hard because the answer depends on relations between agents.

\begin{figure}[t]
    \centering
    \includegraphics[width=0.7\linewidth]{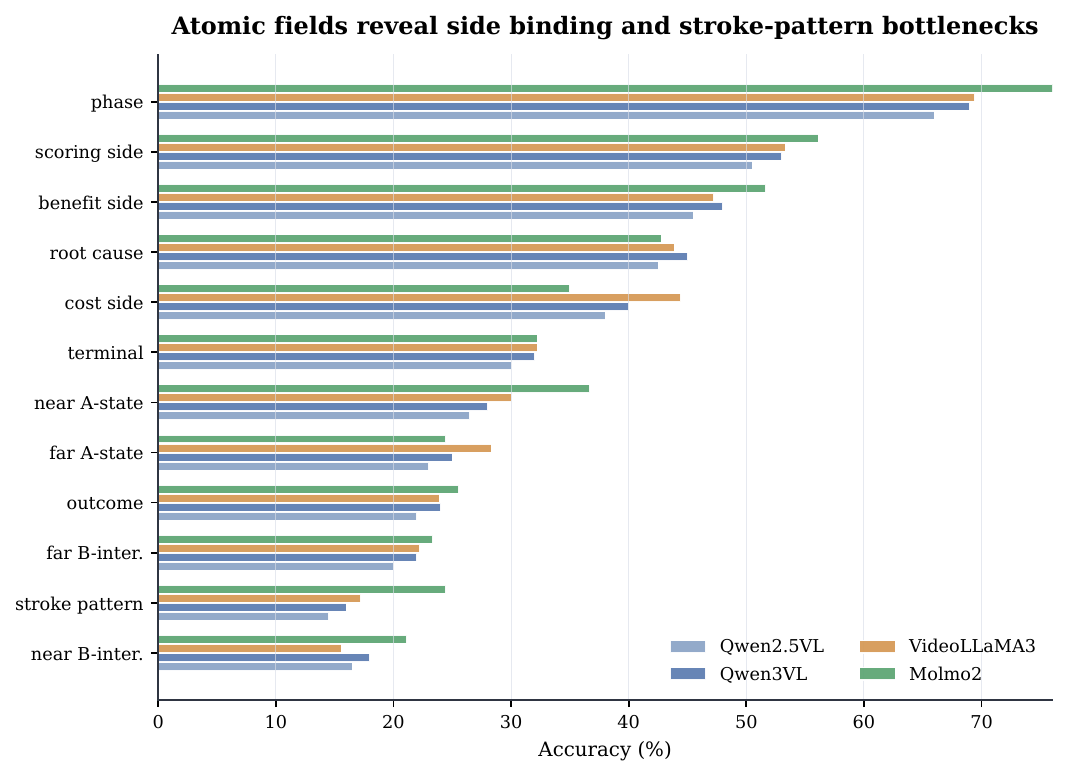}
    \caption{\textbf{L1 field-level diagnosis.} Even within atomic recognition, not all fields are equally easy. Phase and scoring side are more reliable, while near/far interaction fields and stroke pattern remain difficult.}
    \label{fig:l1_field}
\end{figure}

Figure~\ref{fig:l1_field} gives a more fine-grained view of L1. Phase context and scoring side are comparatively stable, but interaction fields are much weaker. Stroke pattern is the hardest L1 field for all models. This is important because later causal questions often depend on these same fields. If the model misreads the terminal stroke or binds the wrong player to the wrong interaction, higher-level reasoning has little chance to be correct.

The field-level results explain why aggregate accuracy alone is insufficient. A model may select or generate a plausible answer locally while still relying on an unstable tactical reading of the scoring moment. These errors are concentrated in the same places highlighted above: spatial state, interaction binding, and terminal evidence. This motivates TacticCheck, a lightweight test-time reranking method that does not try to teach the model new badminton knowledge, but instead checks whether each candidate answer is consistent with the model's own lower-level tactical reading.

\subsection{Improved Performance with Test-Time Consistency Checking}
\label{sec:tacticcheck_perf}
Following Section~\ref{sec:correction}, TacticCheck first uses the model's own L1 predictions as a tactical checklist, including phase, team structure, interaction cues, and score evidence. For higher-level single-choice questions, it reranks candidate options by penalizing options whose tactical signatures conflict with this checklist. The method uses no ground-truth labels during reranking and does not update model parameters.

\begin{table}[t]
\centering
\small
\setlength{\tabcolsep}{4.5pt}
\renewcommand{\arraystretch}{1.12}
\caption{\textbf{Test-time improvement with TacticCheck (\%).} TacticCheck improves all models, but also introduces a smaller number of right-to-wrong flips. This confirms that consistency checking is useful but not error-free.}
\begin{tabular}{lccccc}
\toprule
\textbf{Model} & \textbf{Base} & \textbf{TacticCheck} & \textbf{$\Delta$} & \textbf{W$\rightarrow$R} & \textbf{R$\rightarrow$W} \\
\midrule
Qwen2.5VL   & 27.57 & 35.20 & +7.63 & 12.41 & 4.78 \\
Qwen3VL     & 29.00 & 37.20 & +8.20 & 13.46 & 5.26 \\
VideoLLaMA3 & 32.30 & 39.43 & +7.13 & 12.84 & 5.71 \\
Molmo2      & 31.31 & 39.14 & +7.83 & 12.97 & 5.14 \\
\bottomrule
\end{tabular}

\label{tab:tacticcheck_overall}
\end{table}

\vspace{-5pt}
As shown in Table~\ref{tab:tacticcheck_overall}, TacticCheck improves overall accuracy by $6.55$--$7.83$ percentage points across the four models. The gain is consistent, but we avoid interpreting it as a complete solution. First, TacticCheck accuracy remains below $40\%$ for all models. Second, the method changes some originally correct answers into incorrect ones. This is expected for a test-time consistency reranker: when the model's own checklist is wrong, reranking can propagate the error. The result is therefore best understood as evidence that explicit tactical consistency helps, not that it solves multi-agent tactical reasoning.

These results support the overall story of the benchmark. DoublesEval reveals that the main weakness is not only recognizing badminton actions, but binding visible evidence into a coherent multi-agent tactical chain. A lightweight consistency reranker can recover part of this gap, but the remaining errors show that future VLMs still need stronger interaction modeling rather than only larger visual encoders or better prompt formatting.

\section{Conclusion}
In this work, we introduced DoublesEval, a diagnostic framework that exposes multi-agent tactical reasoning limitations in VLMs through structured evaluation on professional doubles badminton. Our experiments reveal consistent bottlenecks in spatial state recognition, interaction binding, and causal reasoning across four diagnostic levels. 



\bibliography{egbib}
\end{document}